\documentclass[conference]{IEEEtran}
\IEEEoverridecommandlockouts
\usepackage[utf8]{inputenc}
\usepackage[T1]{fontenc}
\usepackage{cite}
\usepackage{microtype}
\usepackage{amsmath,amssymb}
\usepackage{graphicx}
\usepackage{booktabs}
\usepackage{array}
\usepackage{xcolor}
\usepackage{url}
\usepackage[hidelinks]{hyperref}
\hypersetup{pdftitle={}, pdfauthor={}, pdfsubject={}, pdfkeywords={}, pdfcreator={}, pdfproducer={}}

\newcommand{\cm}[1]{\unskip}
\newcommand{\pkglink}{https://osf.io/mqt4h/overview?view_only=96b787a9356049aab0dca50e614eb385}
\newcommand{\Opt}{\mathrm{Opt}}
\newcommand{\val}{\mathrm{val}}
\newcommand{\rqanswer}[2]{%
  \par\addvspace{4pt}%
  \begingroup
  \setlength{\fboxsep}{2pt}%
  \nointerlineskip\noindent
  \fbox{\parbox{\dimexpr\columnwidth-2\fboxsep-2\fboxrule\relax}{%
    \strut\textbf{Answer to #1.} #2\strut}}%
  \par\endgroup\addvspace{4pt}%
}

\newtheorem{proposition}{Proposition}

\begin{document}
\bstctlcite{IEEEcontrol:authors}
\hyphenpenalty=200

\title{Solver-Admitted Metamorphic Testing of\\ Learned Network Optimization Surrogates}

\author{\IEEEauthorblockN{Vineet Jagadeesan Nair\textsuperscript{*}, Stephen Peng\textsuperscript{*}, Arun Ravindran}
\IEEEauthorblockA{AI Science Institute, Boston Consulting Group}
\thanks{\textsuperscript{*}Equal contribution.}}

\maketitle

\begin{abstract}
Equivalent encodings of an optimization problem can produce different learned predictions. Interpreting these differences requires accounting for numerical error and multiple optima. We construct metamorphic tests for optimization solver surrogates with explicit eligibility conditions, output alignment, and numerical thresholds. Reference-solver agreement and independent constraint checks determine which sampled pairs enter the audit. We study alternating-current optimal power flow (AC-OPF), the traveling salesman problem (TSP), the capacitated vehicle routing problem (CVRP), and maximum independent set (MIS). Across five grids, GridSFM-Open violates each of 13 relations on at least one grid, and 12 by more than twice the threshold. Randomly initialized copies of the grid models fail nearly the same relations, so the failures do not require training. Source patches and input canonicalization remove selected discrepancies, but no intervention establishes preserved accuracy and feasibility within the declared margin on every grid. In routing, Policy Optimization with Multiple Optima (POMO) has a single-pass rotation/reflection violation rate that rises from 12.7\% to 89.5\% between 20 and 100 cities. Graph-Based Diffusion Solvers for Combinatorial Optimization (DIFUSCO) violates the MIS disjoint-union relation on 35.6\% of draws. Post hoc normalization within connected components removes these violations on the tested graphs. These findings support reporting metamorphic consistency alongside optimality, feasibility, and accuracy, and validating repair quality separately from consistency.
\end{abstract}


\section{Introduction}

Learned optimization surrogates predict approximate solutions or initialize conventional solvers. Grid foundation models predict generator dispatch for alternating-current optimal power flow (AC-OPF), which meets demand at minimum cost under electrical and thermal constraints \cite{bhanetal2026gridsfm,jinetal2026lumina,puechetal2026genco}. Neural solvers also construct traveling salesman problem (TSP) tours, capacitated vehicle routing problem (CVRP) routes \cite{kwon2020pomo}, and maximum independent sets (MIS) of nonadjacent vertices \cite{sun2023difusco}. Checking every prediction against a reference solve is expensive, creating a test-oracle problem \cite{coffrinetal2018powermodels,wachterbiegler2005implementation,applegate2006tsp,weyuker1982nontestable,barr2015oracle}.

A complementary check asks whether predictions depend on how the same problem is encoded. Renumbering buses, i.e., the grid's connection points, should only renumber predicted settings, while reordering line-table rows should leave them unchanged. Ordinary accuracy and feasibility checks do not directly test this internal consistency. Yet renumbering the standard 14-bus grid (IEEE case14) changes GridSFM-Open's reactive-power prediction by 0.30 per unit (p.u.), or 30 megavolt-amperes reactive (MVAr), about 30,000 times our threshold\cm{RP2}. Per-unit values express quantities relative to their declared bases. Reordering line-table rows changes the prediction by 0.60\,p.u.\cm{RP1} Source patches and a canonical row order remove these discrepancies with frozen weights (Section~\ref{sec:repair}). Both relations also fail within the model's stated range, on grids with at least 500 buses (Fig.~\ref{fig:detection}).

\emph{Metamorphic testing} specifies how outputs should correspond after an input transformation \cite{chen1998metamorphic,chenetal2018metamorphic,seguraetal2016survey}. We derive this correspondence from the optimization problem. A \emph{pull-back} maps transformed predictions to the original coordinates: after renumbering buses, we undo the numbering before comparing generator outputs. Two safeguards make this comparison interpretable. \emph{Admission} checks transformation eligibility and reference-solver agreement. \emph{Calibration} checks whether a numerical threshold, or \emph{floor}, separates noise or round-off from larger discrepancies across the complete process of transforming inputs, predicting, and aligning outputs. Even above that floor, different predictions may be equally correct optima. Propositions~\ref{prop:violation}--\ref{prop:bound} state when a discrepancy implies solution error. Reference solves thus validate the test pairs, while metamorphic comparisons test consistency and help locate faults. Our contributions are:
\begin{itemize}
\item We derive problem-preserving relations, including component additions and splits, and state when violations imply solution error.
\item We audit grid, routing, and MIS models using random weights, controls that isolate specific operations, and vendor-code checks.
\item We localize failures and test how interventions affect consistency, accuracy, and feasibility.
\end{itemize}

\paragraph*{Research questions (RQs)}
\begin{description}
\setlength{\itemsep}{2pt}
\setlength{\parsep}{0pt}
\item[RQ1:] Which sampled transformations pass admission, and can calibration of the complete pipeline distinguish round-off from structural discrepancies? (Section~\ref{sec:oracle})
\item[RQ2:] Which relations do trained grid models violate, compared with random weights and mechanism controls? (Section~\ref{sec:casestudy})
\item[RQ3:] Which mechanisms explain violations, which interventions remove them, and how do those interventions affect accuracy and feasibility? (Section~\ref{sec:repair})
\item[RQ4:] Which relations detect the identified mechanisms, and what smallest sets cover them? (Section~\ref{sec:adequacy})
\item[RQ5:] How do construction, admission, and diagnosis transfer to TSP, CVRP, and MIS? (Section~\ref{sec:routing})
\end{description}

\section{Background and Related Work}\label{sec:related}

\paragraph*{Metamorphic testing of learned systems}
Metamorphic relations expose faults in classifiers \cite{xie2011testing}, driving perception \cite{tian2018deeptest,zhang2018deeproad}, and trajectory prediction with non-unique labels \cite{spiekeretal2024trajectory}. Surveys document their use across learned systems \cite{zhang2022mltesting,riccio2020testing}. Studies of classifier implementation bugs \cite{dwarakanath2018identifying} most closely match our fault-localization work. Optimization adds the need to preserve the underlying problem and distinguish numerical variation, inconsistency, and solution error. Dynasto similarly checks test validity before interpreting outcomes by excluding driving failures caused by implausible adversaries \cite{humeniuketal2026dynasto}.

\paragraph*{Validity-aware scientific testing}
Li et al. combine validity gates, numerical floors, relation specifications, and inserted faults for scientific surrogates, with an analytical floor for a specified operator and mesh \cite{lietal2026domainvaliditygated}. We add mappings between optimal solutions, analysis of non-unique optima, and interventions in released model code. NOETHER derives relations from operator algebras \cite{lietal2026noether}, while MetaTrimmer infers applicability constraints from data and inspection \cite{duquetorres2023metatrimmer}. Our relations and eligibility conditions follow optimization-preserving reductions. MAgg aggregates predictions for Decision TSP, Boolean satisfiability (SAT), and graph edit distance \cite{magg2024}. We test the tours produced by solvers, using reference-solver agreement to check each sampled transformation.

\paragraph*{Equivariance}
An equivariant model transforms its output consistently with its input. The Lie derivative measures departures from this behavior under continuous transformations \cite{gruver2023lie}. Permutation-equivariant layers \cite{maronetal2018invariant,zaheeretal2017deep,battagliaetal2018relational} motivate relabeling consistency, but directed edges and positional features can break it. For example, graph eigenvectors can change sign or basis under relabeling, a known problem with proposed remedies \cite{limetal2023sign,huangetal2024stability}. A deterministic equivariant map also cannot select an optimum unless some optimum is fixed by the input's symmetries \cite{kaba2023symmetry}. Learned canonicalization \cite{kabaetal2023equivariance} and frame averaging \cite{punyetal2022frame} provide broader approaches to choosing a common encoding or averaging over encodings. We test simpler versions for LUMINA-2M (Section~\ref{sec:mechanisms}), as well as discrete transformations that change the number of variables.

\paragraph*{Learned network optimization}
Grid-model evaluations use OPFData \cite{lovettetal2024opfdata} over the Power Grid Library for Optimal Power Flow (PGLib-OPF) grids \cite{babaeinejadsarookolaeeetal2021power} and measure accuracy, feasibility, and outage robustness \cite{pilotoetal2024canos,liuetal2023topologyaware,arowolocremer2026generalization}. Unlike outages, our relations preserve optimal solutions under a reduction. Related grid-foundation work studies topology transfer for power flow and state estimation, rather than AC-OPF \cite{papaioannouetal2026mxgps}. Worst-case verification bounds errors or constraint violations over input regions for direct-current (DC-) and AC-OPF models \cite{venzke2020learning,nellikkathchatzivasileiadis2022physicsinformed}. Our tests yield sampled counterexamples. Routing and MIS evaluations measure gaps from reference solutions \cite{kwon2020pomo,kool2019attention,sun2023difusco}, and a prior non-archival workshop study documents AC-OPF representation dependence \cite{pengnair2026metamorphic}.

\section{Deriving Problem-Preserving Relations}\label{sec:relations}

\subsection{Encodings, Problems and Relations}\label{sec:map}

An \emph{encoding} $d$ stores an instance's node and edge tables, identifiers, row positions, orientations, and other attributes. The \emph{data-to-problem map} $P(d)=(F(d),J(d),\mathrm{out}(d))$ specifies how a solver interprets these data $d$: the feasible set $F$, objective function $J$, and output coordinates $\mathrm{out}$. Let $\Opt(d)$ be the set of global optima, $\val(d)$ their objective value, and $f(d)$ the surrogate prediction. For stochastic models, we include the realized random numbers in $d$ to make each paired computation deterministic. A \emph{sufficient statistic} $s(d)$ retains everything needed to reconstruct $P(d)$. Fields ignored by $P$ are \emph{inert}. For example, electrical data can be represented independently of line-row order. Some outputs also depend on an arbitrary reference, or \emph{gauge}: for e.g., shifting all voltage angles in a connected grid component preserves physical flows. The statistic $s_{\mathrm{phys}}$ omits this reference choice.

\begin{figure}[t]
\centering
\includegraphics[width=\columnwidth]{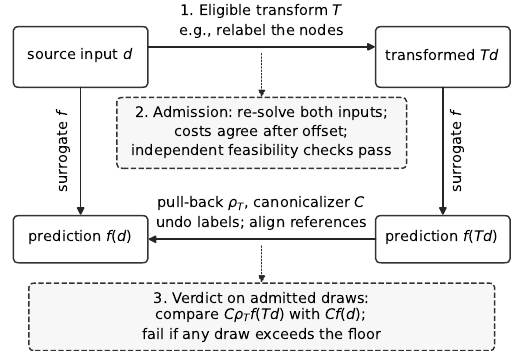}
\caption{From a relation to a test: admission re-solves both inputs to check the sampled transformation, and the verdict compares aligned predictions with the floor. Admission evidence depends on the solver's guarantees.}
\label{fig:recipe}
\end{figure}

A relation $(E_T,T,\rho_T)$ specifies eligibility, an input transformation, and output correspondence. Code checks the eligibility predicate $E_T$ before applying transformation $T$. The \emph{pull-back} $\rho_T:\mathrm{out}(Td)\to\mathrm{out}(d)$ maps outputs to the original coordinates, for example by undoing a permutation or discarding an added subproblem. A canonicalizer $C$ then aligns arbitrary references. Pairing a source instance with a sampled transform gives one \emph{draw} (Fig.~\ref{fig:recipe}). A relation is \emph{valid} if every eligible transformed optimum pulls back to an original optimum after reference correction, with
\[
J(Td)(x')=J(d)(\rho_T x')+\kappa_T(d),\quad \forall x'\in\Opt(Td).
\]
The known objective offset $\kappa_T(d)$ accounts for changes such as adding an independent subproblem or editing constant costs. In our catalogs, it is the added subproblem's optimum for disjoint union, $+1$ for MIS isolated-vertex padding, the summed change for constant-cost edits, and zero otherwise. A \emph{refinement} changes structure, possibly adding variables, while preserving optima under this reduction.

\subsection{Generation Rule and Evidentiary Tiers}\label{sec:tiers}

We construct relations by tracing how stored fields enter $P$. Ignored fields suggest inert edits, while labels, row order, and orientation suggest alternative encodings. Aggregates allow component splitting that preserves their total effect. Padding and insertion require exact reductions, and independent instances allow disjoint union. Each candidate needs eligibility conditions, an output pull-back, an objective offset, and a tier describing the kind of preservation claimed.

\emph{Tier A} preserves the physical problem through numbering, row order, orientation, reference choice, or Euclidean rotation/reflection. \emph{Tier B} changes structure through splitting, padding, union, or insertion. Its pull-back may be many-to-one, but every transformed optimum must reduce to an original optimum. \emph{Tier C} rescales inputs, potentially moving them outside the training distribution or losing equivalence under rounding. We exclude it from the main findings. \emph{Tier D} edits inert fields or objective constants without changing optimal solutions. We verify inertness on the problem actually constructed by the solver interface (Section~\ref{sec:interface}). Tiers describe the relation and its justification, not the severity of a violation.

\section{Instantiating the Recipe}\label{sec:instances}

\subsection{AC Optimal Power Flow}\label{sec:catalogue}

In AC-OPF, \emph{buses} are nodes, \emph{branches} are lines or transformers, and \emph{islands} are connected components. Transformers adjust voltage magnitude through a tap ratio $\tau$ and angle through a phase shift. A voltage-controlled generator bus is called PV, for active power $P$ and voltage magnitude $V$. One slack bus per island fixes the zero angle reference. An OPFData scenario contains component tables, bus links, and context. Loads consume power, and shunts encode local current response to voltage. The variables are the complex voltage $V_i$ at each bus $i$ and each generator $k$'s active and reactive outputs $(p_{g,k},q_{g,k})$. Active power supplies net energy, while reactive power supports voltage. Constraints enforce bus power balance, voltage and dispatch bounds, branch angle bounds, and thermal limits at both branch ends. The generation cost is $J=\sum_k(c_{2,k}p_{g,k}^2+c_{1,k}p_{g,k}+c_{0,k})$. Output channels are voltage magnitude $v_m$, voltage angle $\theta$, and powers $p_g,q_g$.

The sufficient statistic retains bus voltage bounds, aggregate demand and shunt, generator locations, bounds, costs, and island references. For each branch, it retains endpoints, terminal admittance (which maps voltages to currents), rating, and oriented angle bounds. Branches remain separate because their individual limits can bind. For example, parallel branches with ratings 1 and 3 cannot generally be replaced by one branch rated 4. The data converter overwrites stored base voltage, making that field inert here. The canonicalizer $C$ re-anchors angles at the source reference bus.

Table~\ref{tab:catalogue} gives the resulting relations. \emph{R3} reverses branches with tap $=1$ and shift $=0$ exactly, while \emph{tap-side-flip} handles all other branches. A tolerance band could admit a near-unity transformer to \textit{R3} without the required reparameterization, so it cannot uniformly bound electrical error. Reference moves, including \emph{island-ref-mix}, require one reference per island, generators at both old and new references, a PV target, and an eligible inverse. Generator splitting requires $c_2\ge0$ for convexity, satisfied by all scenarios on the five grids studied here.

\begin{table*}[t]
\caption{AC-OPF relations, plus the identity-null probe and the inert-field family. Tier A$\circ$B composes tier-A moves with a tier-B union.}
\label{tab:catalogue}
\centering\footnotesize
\renewcommand{\arraystretch}{0.96}
\setlength{\tabcolsep}{4pt}
\begin{tabular}{@{}l c >{\raggedright\arraybackslash}p{7.25cm} >{\raggedright\arraybackslash}p{2.95cm} >{\raggedright\arraybackslash}p{2.95cm}@{}}
\toprule
Relation & Tier & Transform & Pull-back & Eligibility \\
\midrule
R0 component labels & A & permute generator, load and shunt identifiers & inverse relabeling & always \\
R1 bus labels & A & permute bus identifiers and remap every incidence & inverse relabeling & always \\
R2 row order & A & reorder the rows of an edge table & identity & always \\
R3 branch direction & A & swap a branch's ends and charging, negate and swap angle bounds & identity on nodes & tap $=1$, shift $=0$ exactly \\
tap-side-flip & A & refer a transformer to its other side ($\tau \mapsto 1/\tau$, impedance and charging rescaled) & identity on nodes & complement of R3's domain \\
ref-bus-move & A & move the angle reference within an island & re-anchor at the original & see reference rules \\
component-split & B & split a load or generator 0.5/0.5, bounds halved; $(c_2, c_1, c_0) \mapsto (2c_2, c_1, c_0/2)$ & sum power outputs & $c_2 \ge 0$ for generators \\
parallel-split & B & two parallel halves of a branch (impedance doubled, charging and rating halved) & identity on nodes & at least one branch \\
island-union & B & add a disjoint distractor island $d_D$ with its own reference & drop it; offset $\val(d_D)$ & valid distractor scenario \\
load-as-gen & B & replace a load by a zero-cost generator pinned at $(-p_d, -q_d)$ & drop the added generator & nonzero demand \\
null-pad & B & add a zero load, a zero shunt or a generator pinned at zero & drop the padded elements & always \\
island-ref-mix & A$\circ$B & island-union with permuted islands and moved references & the composed map & union and ref.\ rules \\
basekv-scale & D & scale the base voltage by 0.5, 2 or 10 & identity & always \\
\midrule
identity-null & -- & JavaScript Object Notation (JSON) round trip, key order, $1 \leftrightarrow 1.0$, $0.0 \leftrightarrow -0.0$ & identity & always \\
inert/offset fields & D & perturb $p_g$, $q_g$, $v_g$ set-points, machine base, spare ratings, $c_0$ & identity (offset $\sum \Delta c_0$) & ignored or constant-only \\
\bottomrule
\end{tabular}
\end{table*}

\subsection{Routing and Independent Sets}\label{sec:tsp-recipe}

The Euclidean TSP finds the shortest closed tour through every city. We place cities on the integer grid $[0,10^6]^2$ and give models coordinates scaled to the unit square. The Concorde solver rounds each Euclidean edge length to the nearest integer (EUC\_2D), so the objective is integer tour length. The sufficient statistic is the integer distance matrix up to relabeling. The square's eight rotations and reflections preserve this matrix. CVRP finds depot-to-depot routes serving every customer within vehicle capacity. Its \emph{twin} relation adds a zero-demand copy of a customer. For MIS, the statistic is the graph up to isomorphism, meaning vertex labels are irrelevant. Its structural relations follow two properties: every maximum independent set contains an added isolated vertex, and maximum sets of disjoint components combine into a maximum set of their union (Table~\ref{tab:routing}).

A routing insertion's pull-back deletes the added node and reconnects its neighbors. \emph{M3} (coincident city), \emph{M4} (midpoint insertion), and \emph{twin} require a per-draw \emph{contraction certificate}: deleting that node from any transformed optimum must recover an original optimum. Integer rounding can invalidate this reduction even when the geometric construction appears sound. Section~\ref{sec:admission} explains the checks and the guarantees supplied by Concorde, Hybrid Genetic Search for CVRP (HGS-CVRP), and the constraint-programming SAT solver (CP-SAT). An $n$-city POMO checkpoint receives $n+1$ nodes, while coincident points, zero demands, and isolated MIS vertices are absent from training. These tests therefore probe consistency on equivalent but out-of-distribution instances.

\begin{table}[t]
\caption{Routing and MIS relations. M3, M4, and twin require contraction certificates (Section~\ref{sec:admission}). MIS pad and union increase the optimal set size by \(1\) and by the maximum independent-set size of \(H\), respectively.}
\label{tab:routing}
\centering\footnotesize
\renewcommand{\arraystretch}{0.96}
\setlength{\tabcolsep}{3pt}
\begin{tabular}{@{}>{\raggedright\arraybackslash}p{1.45cm} c >{\raggedright\arraybackslash}p{4.2cm} >{\raggedright\arraybackslash}p{2cm}@{}}
\toprule
Relation & Tier & Transform & Pull-back \\
\midrule
\multicolumn{4}{@{}l}{\emph{TSP: Concorde, certified optimum}} \\
M1 & A & permute the cities & relabel the tour \\
M2 & A & rotate or reflect the square & identity \\
M3 & B & add a coincident copy of a city & drop the copy \\
M4 & B & add the rounded midpoint of an optimal-tour edge & drop the midpoint \\
M5 & C & scale coordinates by 0.5 or 0.25 & identity \\
\midrule
\multicolumn{4}{@{}l}{\emph{CVRP: HGS-CVRP, heuristic agreement}} \\
relabel & A & permute customers, depot fixed & relabel the routes \\
rotate\slash reflect & A & as M2, depot included & identity \\
twin & B & add a zero-demand customer copy & drop copy and empty routes \\
scale & C & scale demand and capacity by 2 or 3 & identity \\
\midrule
\multicolumn{4}{@{}l}{\emph{MIS: CP-SAT, certified optimum}} \\
relabel & A & permute the vertices & relabel the set \\
pad & B & add an isolated vertex & drop it \\
union & B & add a disjoint test graph $H$ & split by part \\
\bottomrule
\end{tabular}
\end{table}

\subsection{What the Catalogs Guarantee}\label{sec:props}

The \href{\pkglink}{\textcolor{blue}{\textbf{software package}}} contains proofs under the stated eligibility conditions and example-based implementation checks.

\begin{proposition}[Validity]\label{prop:validity}
Each relation in Table~\ref{tab:catalogue} is valid on its eligible domain. Tier A preserves the problem up to relabeling and reference choice. Tier B either preserves per-bus totals, as in load component-split and load or shunt null-pad, or follows an exact reduction. These reductions use convexity for generator \textit{component-split}, proportional branch parameters for \textit{parallel-split}, independent components for \textit{island-union}, and fixed dispatch for generator \textit{null-pad} and \textit{load-as-gen}.
\end{proposition} 

Every relation in Table~\ref{tab:catalogue} also preserves all original optima: $\rho_T(\Opt(Td))=\Opt(d)$. In Table~\ref{tab:routing}, tier-A moves preserve integer distances or graph edges, routing refinements (tier B) require contraction certificates, and MIS \textit{pad} and \textit{union} follow from the independent-set properties above.

\paragraph*{Composition and implementation checks}
Eligible AC-OPF tier-A moves form a finite group over reachable encodings in exact arithmetic: they can be composed and undone. Tier-B pull-backs use binary matrices to identify which transformed components contribute to each original component. Composition requires matching dimensions and domains, and inversion requires bijectivity. Powers are summed, while nonadditive quantities such as voltage require one contributor. Implementation checks confirm exact \textit{R3} inversion and \textit{identity-null} preservation. The \textit{tap-side-flip} calculation incurs rounding.

\begin{proposition}[Generation]\label{prop:generation}
The rule in Section~\ref{sec:tiers} generates the relations in Table~\ref{tab:catalogue} from the sufficient statistic $s$. Tier-A moves connect encodings of one physical instance that agree on inert fields, branch families (line/transformer tables), per-bus load/shunt rows, and generator-bus PV/reference labels, provided every transformer has resistance $r>0$ and $|\mathrm{shift}|<\pi$. We make no completeness claim for tier B.
\end{proposition}

\begin{proposition}[What a violation proves]\label{prop:violation}
For a valid admitted draw, a discrepancy beyond round-off shows dependence on the encoding of one physical instance (tier A), inconsistent aligned predictions for problems linked by a reduction (B), or sensitivity to inert fields or objective constants (D). The numerical floor tests whether the discrepancy exceeds round-off. A violation alone does not prove that either solution is wrong, because different optima can both be correct.
\end{proposition}

\begin{proposition}[Error bound]\label{prop:bound}
Let $C$ be linear and let $\delta$ measure $Cf(d)-C\rho_Tf(Td)$ by the largest channel mean or largest componentwise absolute discrepancy (Section~\ref{sec:floor}). If the distance between any two aligned optima in $C(\Opt(d))$ is at most $D$ in that statistic, one aligned prediction is at least $(\delta-D)/2$ away from every aligned optimum. For one component, replace $D$ by that component's range over the optima. For objective discrepancies between feasible answers, the worse answer is at least $\delta$ away from the optimum in the same normalization. This applies to TSP and CVRP, and separately to each MIS component.
\end{proposition}

\begin{IEEEproof}
Both discrepancy statistics are seminorms and obey the triangle inequality. Let $e_1,e_2$ be the aligned predictions' distances to their nearest points $x_1,x_2$ in $C(\Opt(d))$. The distance between $x_1$ and $x_2$ is at most $D$, so $\delta\le e_1+D+e_2$ and $\max(e_1,e_2)\ge(\delta-D)/2$. Feasible objective values lie on the same side of the optimum, so the worse value is at least the pair's difference away from the optimum.
\end{IEEEproof}

The discrepancy must exceed variation among correct answers to yield a positive bound. A unique aligned optimum gives $D=0$ and error at least $\delta/2$. Recovered local solutions supply only a conditional estimate of $D$, since their spread need not enclose every global optimum. A convex relaxation enlarges the feasible set. For minimization, exact coordinate ranges over a relaxation of $\{x\text{ feasible}:J(x)\le J_{\mathrm{ref}}\}$ therefore enclose all global optima: even a local feasible reference costs at least as much as a global optimum. Section~\ref{sec:nonunique} distinguishes this guarantee from our numerical ranges.

\section{Oracle Construction (RQ1)}\label{sec:oracle}

These oracle checks apply to all four domains. Sections~\ref{sec:subjects} and~\ref{sec:routing-subjects} give model and sampling protocols.

\subsection{Admission by Re-Solving}\label{sec:admission}

A mathematically valid relation can be implemented incorrectly or lose validity under integer rounding. We therefore admit a draw only when both reference solves succeed, objectives agree after accounting for the offset, and independent constraint checks pass. Non-finite results are rejected, while missing or ambiguous records remain \emph{unverified}. These admission rules were iterated upon during experimentation stages and then fixed for final runs.

Counts use three units: a \emph{source} instance, a \emph{draw} pairing it with a transform, and a \emph{model draw} evaluating that pair with one model. For AC-OPF, a \emph{cell} groups draws for one model, grid, and relation within a test screen. We run separate screens for the main results (the \emph{headline} screen), numerical calibration, and additional coverage. Pooled counts retain these screen identifiers (Section~\ref{sec:subjects}).

\paragraph*{AC-OPF}
PowerModels.jl and the Interior Point OPTimizer (IPOPT) solve both encodings in AC polar form, with tolerance $10^{-8}$ and a flat start for all variables. Admission requires convergence, relative objective gap
$|J_{\mathrm{ref},T}-(J_{\mathrm{ref}}+\kappa_T)|/\max(|J_{\mathrm{ref}}+\kappa_T|,10^{-12})\le10^{-9}$,
and independently recomputed equality and inequality residuals at most $10^{-6}$. Certificates keyed by input content link each draw to its admission record. We retain 6,370 of 6,385 distinct draws, corresponding to 105,185 of 105,364 model draws, with none unverified\cm{DV1}\cm{DV2}. All rejections are case14 generator splits at binding bounds, where IPOPT solver variability exceeds the objective tolerance. Excluding them changes no AC-OPF cell verdict\cm{DV3a}.

\paragraph*{TSP}
Concorde \cite{applegate2006tsp} certifies both optima, but admission also requires the pulled-back optimum to retain the original length and every transformed optimum to contract correctly. \textit{M1/M2} preserve integer distances. For an inserted node $k$, \textit{M3/M4} require $d(i,j)\le d(i,k)+d(k,j)$ for every pair of other nodes, where $d(i,j)$ is their integer distance. Deleting $k$ then cannot lengthen a tour. \textit{M4} also requires $d(a,k)+d(k,b)=d(a,b)$ on the subdivided optimal-tour edge $(a,b)$, allowing insertion without changing that tour's length. These checks use the solver's integer distance matrix because rounding can break the reduction. All 21,000 Concorde calls certify an optimum\cm{T5}, yet \textit{M3} admits only 998, 975, and 916 of 1,000 draws at $n=20$, 50, and 100, and M4 admits 687, 640, and 578\cm{T6}. Exact solutions to both encoded problems do not by themselves validate the transformations between them.

\paragraph*{CVRP and MIS}
HGS-CVRP \cite{vidal2022hgs} supplies heuristic agreement: source and pulled-back costs must match exactly, and \emph{twin} must pass the contraction certificate. \textit{Relabel}, \textit{rotate/reflect}, and \textit{twin} admit 958, 961, and 685 of 1,000 draws. The certificate causes 286 of \textit{twin}'s 315 rejections\cm{CV2}. HGS-CVRP returns different best costs on 3.9--4.2\% of isomorphic instances, causing those draws to be rejected\cm{CV2b}. For MIS, CP-SAT \cite{ortools_cpsat} proves all 1,250 optima, and every draw is admitted\cm{MIS1}.

\paragraph*{Certificate strength}
Admission evidence depends on the solver. Concorde and CP-SAT certify global optima, HGS-CVRP supplies heuristic agreement, and IPOPT supplies converged local solutions \cite{wachterbiegler2005implementation,bukhshetal2013local}. Variability among local solutions can cause valid tests to be rejected. Conversely, matching objectives cannot exclude every transformation error. The eligibility proofs establish relation validity, while solver agreement checks the sampled implementation.

\subsection{The Numerical Floor}\label{sec:floor}

Equivalent inputs can change the sequence of floating-point operations. For AC-OPF, we therefore declared two floors before the audit: $10^{-6}$ for the \emph{worst-channel mean} and $10^{-5}$ for the \emph{worst component}. The first averages absolute discrepancies within each output channel, then takes the largest mean. The second takes the largest individual discrepancy. Together they detect broad and localized changes in 32-bit arithmetic (float32), measured in p.u.\ for $v_m,p_g,q_g$ and radians for $\theta$. \emph{Identity-null} is exact on 3,456 draws\cm{B1}, checking repeatability but not arithmetic changes caused by transformations. Changing PyTorch threads from one to six gives a worst-channel mean spread of at most $2.24\times10^{-7}$ over 12,096 draws\cm{FG2}.

\begin{figure}[t]
\centering
\includegraphics[width=\columnwidth]{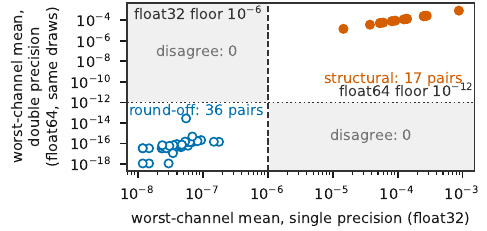}
\caption{Precision calibration separates 36 round-off and 17 structural model--relation pairs. Axes show worst-channel means for the same draws in float32 and float64. Dashed and dotted lines mark the respective floors ($10^{-6}$, $10^{-12}$). The shaded quadrants, where the precision-specific verdicts would disagree, are empty. Five pairs from the float32-only released checkpoints are not plotted.}
\label{fig:floor}
\end{figure}

For 64-bit (float64) calibration, the corresponding mean and component floors are $10^{-12}$ and $10^{-11}$. We pool case14 and case500 by model and relation, selecting float32 worst-channel means in $[10^{-8},10^{-3}]$. Rerunning the same draws gives 53 matched model--relation pairs: 36 pass both precision-specific rules and 17 fail both (Fig.~\ref{fig:floor}). We classify the first group as round-off and the second as structural discrepancies. In float32, the largest round-off value is $1.8\times10^{-7}$ and the smallest structural value is $1.46\times10^{-5}$. Structural values change by at most 0.022\% between precisions\cm{FG1}. On larger grids, relabeling round-off can cross the fixed floor. Precision reruns of the Geometric Neural Corrective Optimizer (GENCO) classify 41 of its 131 failing cells as round-off, excluding them from structural counts\cm{E2}\cm{E6}.

Routing compares lengths rather than node sequences, allowing distinct routes with equal cost. Let $t$ and $t'$ be the source and transformed predictions. We evaluate both on the original integer-distance instance after pull-back: $\delta=|L(\rho_Tt')-L(t)|/L_{\rm ref}$. Here $L$ is total length, and $L_{\rm ref}$ is Concorde's optimum for TSP or HGS-CVRP's cost for CVRP. A $10^{-9}$ floor requires exact equality at these integer scales. MIS likewise compares set sizes rather than vertex identities. For graphs $G,H$ and union prediction $S=f(G\cup H)$, let $\delta_G=\bigl||f(G)|-|S\cap V(G)|\bigr|$ and define $\delta_H$ analogously, where $V(G)$ is the vertex set. We use $\delta_G+\delta_H$ so that a gain in one component cannot cancel a loss in the other. Applying Proposition~\ref{prop:bound} to each component shows that one of $f(G)\cup f(H)$ and $S$ has objective error at least $(\delta_G+\delta_H)/2$.

\subsection{Verdict Rule}\label{sec:verdict}

For admitted AC-OPF draws, let $\delta_c=|C\rho_Tf(Td)-Cf(d)|$ elementwise on channel $c$. A cell fails if any draw has $\delta_{\rm mean}=\max_c\operatorname{mean}(\delta_c)>10^{-6}$ or $\delta_{\max}=\max_c\max(\delta_c)>10^{-5}$. One draw supplies a counterexample. A pass means only that the admitted draws revealed no violation. Five failing cells have a deciding draw within twice the floor, so we flag their evidence as borderline\cm{DV3e}. Routing fails above its $10^{-9}$ floor, and MIS fails on nonzero integer discrepancy. Draws sharing a source are dependent. Grid comparisons therefore use paired cells, while routing/MIS rates describe admitted draws.

The safeguards address different failure modes. Requiring exact equality would flag all 36 round-off pairs\cm{FG1}, while using only the mean would miss 12 component-level failures\cm{DV3b}. Testing more sources and draws overturns two of GENCO's eight unanimous passes\cm{E14}\cm{DD12b}. Admission excludes up to 42.2\% of attempted TSP \textit{M3/M4} draws because they lack the required certificates\cm{T6}.

\subsection{Non-Unique Optima}\label{sec:nonunique}

For AC-OPF, we first check whether different predictions could both be correct optima. This analysis uses an admitted subset of 4,680 paired draws in 520 cells from two released checkpoints and six random-weight runs over five grids. The associated feasibility records contain 7,140 unique predictions, accounting for reused sources and extra transform seeds. None is strictly AC-feasible: every prediction's largest bus-level active or reactive power mismatch exceeds 0.51\,p.u.\cm{NU1} Summing generator powers per bus preserves all 520 verdicts, ruling out cost-free trades between co-located generators as an explanation\cm{NU2}. We then compare discrepancy sizes with the range of solutions recovered by the reference solver.

We run multi-start IPOPT from 27 initial points on 15 scenarios and recover one objective cluster per scenario\cm{NU3}. We form clusters by joining adjacent sorted feasible costs within relative tolerance $10^{-6}$. To estimate solution spread, we retain solutions within a relative $10^{-5}$ of the best recovered cost. Their largest componentwise difference is at most $2.4\times10^{-5}$ ($3.8\times10^{-4}$ on case4661)\cm{NU4}. Counterexamples exceed 100 times that spread in 100 of 106 failing released-model cells\cm{NU5}. The closest cases occur on case4661, including GridSFM \emph{R0} within the spread\cm{NU6}. Using this spread as $D$ in Proposition~\ref{prop:bound} gives positive estimated error bounds in 105 of 106 cells\cm{NU8}, conditional on undiscovered optima not widening the set.

We also estimate ranges intended to enclose the optima using a quadratic convex (QC) relaxation \cite{coffrinetal2018powermodels}. Auxiliary minimization and maximization tighten variable bounds \cite{gopinathetal2020proving}. We then estimate each queried component's range with IPOPT and add $10^{-6}$ to its computed width. Substituting these widths for $D$ gives positive estimated error bounds in 21/21 failing released-model cells on case14 and 18/21 on case118. The gap between the reference cost and relaxed objective, $(J_{\mathrm{ref}}-J_{\mathrm{QC}})/|J_{\mathrm{ref}}|$, is at most $1.6\times10^{-4}$\cm{NU12}. We omitted this bound tightening on larger grids because of its prohibitive cost. Without tightening, 8/21 cells each on case500/case2000 and none on case4661 have positive estimates, at gaps of about 0.2--2\%\cm{NU9}\cm{NU10}. These 55 estimates remain conditional on numerical optimization accuracy: neither the computed extrema nor the added margin certifies enclosure\cm{NU13}. The other 51 cells rely on the multistart estimates alone.

\rqanswer{RQ1}{Admission excludes draws without the required evidence. Calibration separates round-off from structural discrepancies in the 53 matched model--relation pairs. Local-solver agreement and calibrated floors do not certify correctness. Larger GENCO grids need precision reruns, which are unavailable for the float32-only releases.}

\section{Case Study: Power-Grid Foundation Models}\label{sec:casestudy}

\subsection{Subjects, Controls and Protocol}\label{sec:subjects}

We test GridSFM-Open v1.1 \cite{bhanetal2026gridsfm,yangetal2026gridsfm,microsoftgridsfmopen2026modelcard,microsoftgridsfm2026software}, LUMINA-2M v0.1.3 \cite{jinetal2026lumina,argonnelumina2m2026modelcard}, and GENCO Base \cite{puechetal2026genco}. GENCO seeds 42, 3, and 17 use GridFM graphkit's pinned reproduction \cite{gridfmgraphkit2026v081}. Seed 42 matches published case14 metrics within 0.3\%\cm{E1}. Fourteen random-weight \emph{mechanism controls} isolate design choices involving aggregation, stored direction, positional features, index filters, and variable partitions. We check their verdicts in float64\cm{B4}\cm{DV7}. Two controls differ only in their use of raw graph-Laplacian eigenvectors, whose signs or bases can change under relabeling. Controls inspired by Deep Constraint Completion and Correction (DC3) \cite{dontietal2021dc3} select variables by index or physical bus role and complete the rest by averaging. They test partition choice, rather than DC3's solver or training.

We use case14, case118, Grid Optimization Competition case500 and case2000, and OPFData case4661 with N$-$1 perturbations, which remove one branch or generator \cite{lovettetal2024opfdata}. Case labels give bus counts. GENCO excludes case4661, and its random-weight runs and mechanism controls use case14 and case500\cm{DV6}\cm{E9}\cm{DV7}. Each headline cell has three sources and three draws per source. The coverage screen retests passing case14 and case500 cells on ten additional stratified sources with ten draws each\cm{DD6d}. The headline, calibration, and coverage screens contain 831 cells, one unevaluable. GENCO adds 156 cells\cm{DV3}\cm{E2}.

Our adapter pipeline reproduces vendor inference bitwise on all 200 GridSFM and 400 LUMINA rows\cm{E0-G4}\cm{E0-L2}. We also replay the draws that motivated repairs through vendor code\cm{E0-G6}\cm{E0-L3}. Published metrics are less reproducible: GridSFM's case500 cost errors differ between its model card and pinned vendor script\cm{E0-G1}\cm{E0-G2}, while LUMINA provides no reproducible metric\cm{E0-L1}. Our findings concern the \emph{tested integration}, meaning the pinned vendor code run through our adapter\cm{E0-A}.

\begin{figure*}[t]
\centering
\includegraphics[width=\textwidth]{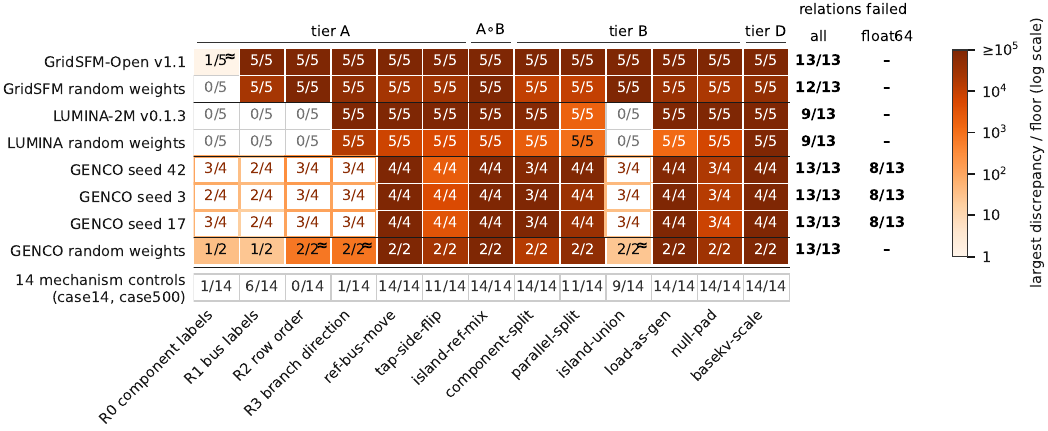}
\caption{Numbers give failing grids over evaluated grids. Color shows the largest $\max(\delta_{\rm mean}/10^{-6},\delta_{\max}/10^{-5})$ on a logarithmic scale capped at $10^5$, with white indicating all pass. Outlined GENCO cells fail only through float32 round-off. The float64 column counts relations that still fail after precision reruns. $\approx$ marks a deciding draw within twice the floor. The bottom row counts mechanism controls failing on case14 or case500.}
\label{fig:detection}
\end{figure*}

\subsection{Detection (RQ2)}\label{sec:detection}

Fig.~\ref{fig:detection} reports 504 failing cells among 831\cm{DV3}. GridSFM fails all 13 relations on at least one grid\cm{DV4}, with twelve exceeding twice the floor. The exception is \textit{R0}: one case4661 draw exceeds the component floor by 19.2\% but lies within the spread of recovered solutions\cm{DV3e}. LUMINA fails nine relations, passing \textit{R0}, \textit{R1}, \textit{R2}, and \textit{island-union} on all five grids\cm{DV4}\cm{DV7}. Each GENCO seed fails all 13 under the float32 rule, but only eight remain structural after precision reruns. Seeds agree on 51 of 52 grid--relation cells\cm{E3}\cm{DD7}. Every trained architecture fails a relation in each tested tier: A, B, and D.

Across 148 failing released-model cells in the headline and calibration screens, we divide the worst discrepancy by the same channel's mean absolute error (MAE). This ratio has median 0.46 and maximum 98\cm{DV3c}. GENCO's structural discrepancies reach a median of 2.7 and maximum of 3,147 times its median MAE\cm{E10}.

\subsection{Trained Versus Random Weights}\label{sec:random}

We repeat the tests with random weights to assess whether failures require training. Random-weight GridSFM matches the released failure pattern except \textit{R0}, and random-weight LUMINA matches it exactly\cm{DV4}\cm{DV7}. Trained GENCO and its random-weight counterpart have the same structural failures, differing only through round-off\cm{E9}\cm{DV3e}. A failure confined to trained weights would establish weight dependence, but would not identify its cause. The mechanism controls help distinguish design choices: index-sensitive features fail \textit{R1}, stored direction fails \textit{R3}, and the index-based partition also fails \textit{R0}. Six relations fail all 14 controls, while \textit{R2} fails none. Those seven relations therefore do not distinguish the tested designs\cm{DV7}.

\subsection{Inert Fields}\label{sec:interface}

PowerModels constructs identical problems for all 540 inert-field and constant-cost edits, apart from the expected objective offset $\sum\Delta c_0$ for the latter\cm{G1d}\cm{G1e}. Positive controls deliberately change ratings and costs, altering the constructed problems on all five grids\cm{G1h}. Despite this equivalence, GridSFM predictions change when spare ratings or machine base change\cm{G1a}, and LUMINA predictions change when $c_0$ changes\cm{G1b}. The LUMINA edits exceed the observed $c_0$ range, so they are stress tests of sensitivity to an objective constant.

\rqanswer{RQ2}{All three architectures violate problem-preserving relations, and random-weight failures show that training is not required. The controls connect failure patterns to design choices that depend on the input representation.}

\section{Fault Localization and Repair Validation}\label{sec:repair}

Changing an operation with fixed model weights helps establish its causal role on tested inputs. If an intervention changes several operations together, the result supports their combined effect without showing that every change is necessary.

\subsection{Protocol}\label{sec:protocol}

We compare aligned features and hidden activations after each of GridSFM's eight blocks. At successive depths, we replace transformed features with aligned features from the source computation. Complete substitution nearly eliminates \emph{R1} discrepancies on case14 and case500, locating their origin in the feature encoder\cm{FG5}. We then patch the responsible operation or canonicalize its inputs and repeat the admitted tests.

Tests with frozen weights use 20 sources per grid and three draws per source on case14, case118, case500, and case2000. Quality checks use 100 paired held-out scenarios per grid (80 on case2000). We measure accuracy by per-channel MAE against the reference solution and feasibility by the maximum power-balance residual and maximum operating-limit violation. For each measure, we divide the variant's mean across scenarios by the released checkpoint's mean. A ratio above one indicates deterioration. We declare \emph{non-inferiority}, meaning that the increase stays within a 10\% margin, only when the one-sided 95\% paired-bootstrap upper limit is at most 1.10.

\subsection{Localized Mechanisms}\label{sec:mechanisms}

Two mechanisms affect sign features. The \emph{cycle sign} mechanism assigns signs to edges along a closed grid path using bus indices rather than stored edge direction. The \emph{R1} draw from the introduction flips 13 of 25 signs. The cycle sign patch uses stored direction, reducing that draw's worst-component discrepancy from 0.296 to $4.8\times10^{-7}$\cm{RP2}. The \emph{flow sign} mechanism turns tiny linearized DC-power flow errors near zero into changes in a sign feature\cm{RP3}. The flow sign patch sets this feature to zero when the computed flow satisfies $|P_{\rm DC}|\le10^{-6}$\,p.u.

The \emph{landmark index} mechanism selects structural reference nodes (landmarks) by bus index. A case500 \emph{R1} draw changes 62 of 64 landmarks\cm{RP4}. The landmark patch ranks candidates within reference-bus, generator-bus, and all-bus pools by weighted degree rounded to eight decimals, then by bus attributes. Exact ties retain index dependence. Comparing subsets of the patches attributes the \emph{R1} discrepancies remaining after the cycle sign patch to flow sign and landmark index.

The \emph{row position} mechanism has three sources in GridSFM: a spanning tree weighted by row order, transformer features read by position from line rows, and an assumption that link row $i$ belongs to component $i$. Sorting every edge table into a canonical order independent of labels removes all three sources of variation, making the introductory \emph{R2} inputs bit-identical\cm{RP5}. LUMINA's \emph{stored direction} mechanism arises because it sends messages only in each branch's stored direction and ignores edge features. Reversing the stored endpoints therefore changes its computation\cm{RP6}.

GridSFM variants are cumulative: \emph{patched} applies the three source patches, \emph{canonical} adds the canonical row order, and \emph{fine-tuned} fine-tunes canonical GridSFM with three seeds. For LUMINA, \emph{averaged} averages predictions for the input and its all-branch reversal. Its \emph{canonical} variant gives every branch a fixed orientation. Canonicalization selects a common encoding while leaving the encoding-sensitive operation in place. The patches, averaged LUMINA, and a lexicographic row sort were predeclared. Patch subsets and the other variants ($\dagger$) were developed post hoc.

\begin{figure}[t]
\centering
\includegraphics[width=\columnwidth]{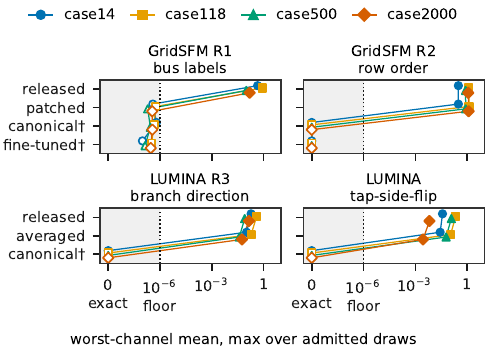}
\caption{Consistency on each target relation by variant, measured by the largest worst-channel mean over 60 admitted draws per grid. Filled markers fail the verdict rule, and leftmost points are exact zeros. GridSFM variants are cumulative. Fine-tuned GridSFM shows the worst of 3 seeds. $\dagger$: post hoc.}
\label{fig:repair-effect}
\end{figure}

\subsection{Patch-and-Retest (RQ3)}\label{sec:patch}

Patched GridSFM reduces \emph{R1}'s worst-channel mean to at most $5.7\times10^{-7}$. Canonical GridSFM keeps it below $6.2\times10^{-7}$ and makes \emph{R2} zero on all four grids (Fig.~\ref{fig:repair-effect})\cm{DD9}. Only one of the four predeclared targets for individual changes is met: the lexicographic row sort removes \emph{R2} failures on all grids. Applied alone, the cycle-sign patch leaves \emph{R3} above the mean floor on some grids, the flow-sign patch leaves \emph{R1} above it on case14, and the landmark patch leaves \emph{R1} above it on the three larger grids\cm{A1}. Every GridSFM variant retains \emph{R3} and \emph{tap-side-flip} discrepancies\cm{DD9}. Canonical LUMINA removes both on all 240 draws per relation\cm{AL6}, whereas averaged LUMINA retains 39--55\% of the original \emph{R3} discrepancy\cm{AL3}.

\begin{figure}[t]
\centering
\includegraphics[width=\columnwidth]{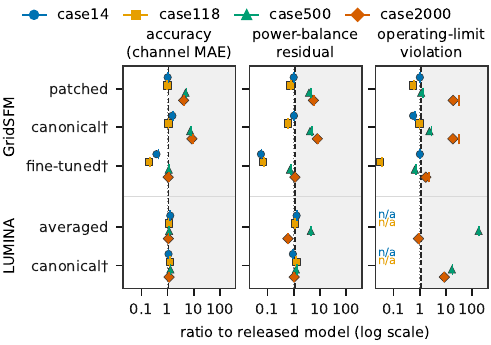}
\caption{Consistency repair does not establish preserved quality everywhere. Markers give ratios of means across scenarios to the released model's means. Caps show one-sided 95\% upper limits, and the dashed line marks the 1.10 non-inferiority margin. Accuracy shows the channel with the largest upper limit per variant and grid. Fine-tuned GridSFM also shows its worst seed. n/a denotes an undefined ratio because the baseline mean is zero. Operating-limit violations use absolute p.u.\cm{FR6}}
\label{fig:repair-cost}
\end{figure}

\begin{figure}[t]
\centering
\includegraphics[width=\columnwidth]{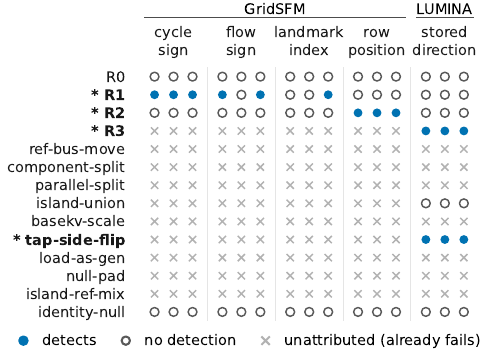}
\caption{Three relations cover the five identified mechanisms. Each column re-enables one mechanism in canonical GridSFM or LUMINA. A filled mark means the relation passes with the repair and fails without it. A $\times$ means the canonical variant already fails. The three marks represent case14, case118, and case500. Stars identify the two minimum covering sets.}
\label{fig:adequacy}
\end{figure}

\begin{figure*}[t]
\centering
\includegraphics[width=\textwidth]{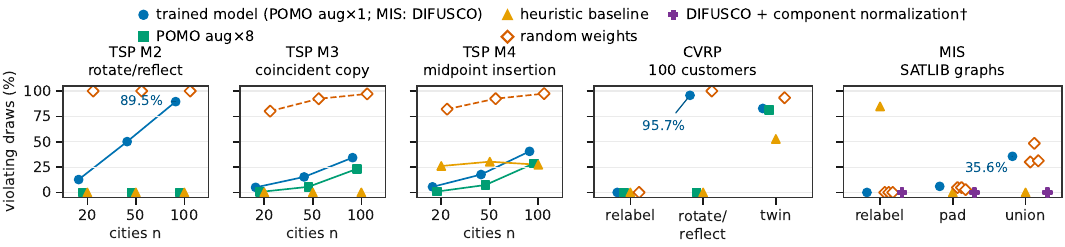}
\caption{Violation rates over admitted draws. Trained models are POMO aug$\times$1 (TSP, CVRP) and DIFUSCO (MIS). Heuristic baselines are nearest neighbor, nearest-feasible neighbor, and greedy minimum degree. Random-weight rates pool three seeds for TSP and CVRP, while MIS shows three seeds. POMO aug$\times$8 removes \textit{rotate/reflect} failures up to float32 ties but leaves \textit{M3}, \textit{M4}, and \textit{twin} failures. Post hoc ($\dagger$) component normalization removes DIFUSCO's \textit{pad} and \textit{union} violations. CVRP references are heuristic.}
\label{fig:transfer}
\end{figure*}

Canonical GridSFM (Fig.~\ref{fig:repair-cost}) establishes accuracy non-inferiority only on case118 and feasibility non-inferiority on case14 and case118. On case2000, upper limits exceed the margin for all three measures\cm{FG4}\cm{A7}\cm{A8}. Canonical LUMINA increases observed angle MAE on the three larger grids, and its case500 operating-limit ratio has an upper limit of 17.8\cm{AL10}\cm{AF5}. These limits reflect uncertainty, so failure to establish non-inferiority does not by itself demonstrate degradation.

Fine-tuned GridSFM keeps \textit{R1} below the floor and R2 exactly zero\cm{AP1}. Its accuracy is non-inferior to released GridSFM in 46 of 48 seed--grid--channel tests\cm{AP4}. When released GridSFM is also fine-tuned by the same procedure, the repaired variant fails to establish non-inferiority in 5--11 of 16 grid--channel tests per seed\cm{AP5}. No tested GridSFM or LUMINA prediction meets both the 10-megawatt (MW) active-power and 10-MVAr reactive-power tolerances at every bus\cm{FR2}\cm{FR3}.

\rqanswer{RQ3}{Interventions support the identified mechanisms and remove selected discrepancies, including after fine-tuning. Quality remains a separate requirement: no variant establishes non-inferiority on every grid and measure.}

\subsection{Mechanism Detection (RQ4)}\label{sec:adequacy}

Starting from canonical GridSFM or canonical LUMINA, we undo one repair at a time while keeping the others in place. A relation detects the re-enabled mechanism only if it passes with the repair and fails without it. The screen uses 20 sources per grid, disjoint from the discovery examples, on case14, case118, and case500, with three draws each. Admission succeeds for 2,095 of 2,102 certificates\cm{HA1}\cm{HA2}. \textit{R1} detects \textit{cycle sign}, \textit{flow sign}, and \textit{landmark index}. \textit{R2} detects \textit{row position}, while \textit{R3} and \textit{tap-side-flip} detect \textit{stored direction}. Fig.~\ref{fig:adequacy} shows coverage by grid\cm{HA4}\cm{HA5}\cm{HA6}\cm{HA7}\cm{HA8}. Canonical GridSFM already fails ten of 13 relations and canonical LUMINA seven, so those failures cannot be attributed to a re-enabled mechanism\cm{HA3}\cm{HA8}. This analysis covers mechanisms discovered using the same relations, rather than an independent sample of faults.

\rqanswer{RQ4}{Pooling evidence across 3 grids, 2 minimum sets detect all 5 discovered mechanisms: $\{$\textit{R1},\textit{R2},\textit{R3}$\}$ and $\{$\textit{R1},\textit{R2},\textit{tap-side-flip}$\}$\cm{HA9}. Their coverage does not extend to arbitrary faults or the remaining unexplained failures.}

\section{Case Study: Neural Routing and MIS Solvers}\label{sec:routing}

\subsection{Subjects and Protocol}\label{sec:routing-subjects}

For RQ5, we apply the same construction, admission, and comparison procedure to routing and MIS. We report \emph{base gaps} to show solution quality before transformation: $(L-L_{\rm ref})/L_{\rm ref}$ for routing and $(\alpha-|S|)/\alpha$ for MIS, where $\alpha$ is the certified maximum set size. CVRP's reference remains heuristic.

POMO \cite{kwon2020pomo} supplies TSP checkpoints for 20, 50, and 100 cities. Its single-pass inference, aug$\times$1, greedily decodes from every start and keeps the best tour. The aug$\times$8 variant also evaluates all eight rotations/reflections of the square, making \textit{M2} hold in exact arithmetic. We measure how single-pass orientation dependence varies with size and whether \textit{M3/M4} failures persist with aug$\times$8. Each size uses 1,000 instances, one draw per relation, three random-weight seeds, and a nearest-neighbor baseline that keeps the best tour across all starts. For CVRP with 100 customers (CVRP100), we use 1,000 training-distribution instances, trained and random-weight models, and a nearest-feasible-neighbor baseline.

DIFUSCO refines noisy vertex choices before decoding an independent set. Its Satisfiability Library (SATLIB) checkpoint \cite{sun2023difusco} has a mean base gap of 0.31\% on our 250 held-out graphs, versus 0.34\% published on all 500\cm{MIS8}. Controls are random weights and a greedy minimum-degree baseline with smallest-index tie-breaking. Stochastic tests must align random choices as well as graphs. DIFUSCO's random-number matrices for initialization and refinement have one column per vertex. \textit{Relabel} permutes columns, \textit{pad} adds an independent column, and \textit{union} concatenates the matrices used for separate $G$ and $H$ predictions. This keeps each vertex's random choices aligned across encodings. Merely reusing a seed changes set size on 36.4\% of relabelings\cm{MIS7}.

\subsection{TSP}\label{sec:tsp-findings}

POMO aug$\times$1 has no \textit{M1 (relabel)} violation in 3,000 certified draws\cm{T10}, but \textit{M2 (rotate/reflect)} violations rise from 12.7\% to 50.1\% to 89.5\% at 20, 50, and 100 cities\cm{T7}. Small mean optimality gaps, respectively 0.023\%, 0.126\%, and 0.348\%, therefore coexist with frequent orientation dependence\cm{T13}. At 100 cities, \textit{M3 (coincident copy)} and \textit{M4 (midpoint insertion)} violation rates reach 34.3\% and 40.5\%\cm{T8}. Even aug$\times$8 fails \textit{M3} on 23\% and \textit{M4} on 29\%\cm{T9} (Fig.~\ref{fig:transfer}). Random-weight POMO fails \textit{M2} on every draw and \textit{M3/M4} on 93.8--99.7\% at $n=100$\cm{T12}, but fails \textit{M1} only on two float32 near-ties\cm{T10}. The \textit{M1} result is consistent with its permutation-equivariant encoder \cite{kool2019attention,kwon2020pomo}, although decoding and tie-breaking limit end-to-end consistency. The nearest-neighbor baseline fails only \textit{M4}, on 26.1--30.3\% of draws, because the midpoint changes its greedy next-city choice\cm{T11}. \textit{M5 (scale)} remains outside the certified findings.

\subsection{CVRP and MIS}\label{sec:cvrp-mis}

On CVRP, POMO aug$\times$1 passes \textit{relabel} on all 958 admitted draws but violates \textit{rotate/reflect} on 95.7\% and \textit{twin} on 82.8\%. The corresponding median discrepancies are 0.35\% and 0.16\% of reference cost, compared with a mean base gap of 1.76\%\cm{CV3}\cm{CV7}. Aug$\times$8 removes only the \textit{rotate/reflect} failures\cm{CV4}. Violation rates remain similar when we drop the HGS-CVRP agreement requirement\cm{CV9}. The \textit{scale} relation leaves model inputs bit-identical and yields no violations\cm{CV8}. The nearest-feasible-neighbor baseline fails only \textit{twin}, on 52.7\% of draws. The added customer requires no capacity and can change the next greedy choice, producing costlier routes in 356 of 361 failures\cm{CV5}. Random weights fail every admitted \textit{rotate/reflect} draw\cm{CV6}, so orientation dependence does not require training.

DIFUSCO passes MIS \textit{relabel} but violates \textit{pad} on 6.0\% and \textit{union} on 35.6\% of 250 draws per relation, with discrepancies of up to three and four vertices, respectively\cm{MIS2}\cm{MIS3}\cm{MIS4}. Random weights likewise pass \textit{relabel} but fail \textit{pad} and \textit{union}\cm{MIS5}. The greedy minimum-degree baseline fails only \textit{relabel}, on 84.8\%, because of its index-based tie-break\cm{MIS6}. DIFUSCO normalizes final scores across all vertices, allowing an added disconnected component to change scores in the original graph. Relabeling preserves this normalization. On three pilot graphs, \textit{union} also changes first-step probabilities more than \textit{relabel} does\cm{MIS9}. We test a post hoc intervention, \emph{component normalization}, that normalizes within each connected component. It leaves all 250 untransformed predictions unchanged and removes the tested \textit{pad} and \textit{union} violations\cm{MIS11}.

\rqanswer{RQ5}{Problem-specific reductions and certificates transfer the recipe across domains. Targeted interventions remove some failures, but the mechanisms behind POMO's \textit{M3}, \textit{M4}, and \textit{twin} failures remain unlocalized, and CVRP admission remains heuristic.}

\section{Threats to Validity}\label{sec:threats}

\paragraph*{Construct validity}
Calibration is empirical and supplies no universal round-off bound. Without precision controls, we interpret the float32-only releases' discrepancies by magnitude. Tier assignment also requires judgment about the applicable reduction. For AC-OPF, admission establishes agreement between local solves. Neither multistart spread nor unvalidated QC extrema certify the range of global optima. Recovering one objective cluster\cm{NU3} cannot exclude undiscovered optima, and bounds on global optima do not apply to costlier local solutions.

\paragraph*{Internal validity}
The interventions support causal explanations on tested inputs, but canonical GridSFM and LUMINA, fine-tuned GridSFM, and DIFUSCO component normalization were developed post hoc. Mechanism detection uses disjoint sources to limit reuse of discovery examples, yet covers only discovered mechanisms, one seed stream, and no case2000 run. Without multiple-comparison control, quality tests give individual confidence statements rather than simultaneous guarantees.

\paragraph*{Reliability}
Screens use macOS arm64, fine-tuning uses a graphics processor, and initial QC computations use Linux x86-64\cm{NU11}. IPOPT's cross-platform controls agree\cm{X1}, but Linux replay changes above-floor magnitudes by up to 53\%. It preserves all tested OPF verdicts and all TSP verdicts except one float32 aug$\times$8 tie at $n=50$\cm{T9}. CVRP and MIS were not replayed.

\paragraph*{External validity}
The selected subjects and grids establish counterexamples rather than failure prevalence among learned solvers. Training coverage differs: all five grids occur in LUMINA's training card, while case14 and case118 lie outside GridSFM's stated range. We have not compared the protocol's cost or effectiveness with ordinary accuracy or residual checks, so its relative efficiency remains unknown.

\section{Conclusion}\label{sec:conclusion}

Solver-admitted metamorphic testing exposes dependence on input representations that ordinary prediction metrics do not directly measure. The grid and MIS investigations locate failures in index use, row positions, stored directions, and normalization across disconnected components. Random-weight controls show that failures can occur before training, while routing results show that frequent violations can coexist with small optimality gaps. Relabeling or symmetry tests alone leave other failures undetected, so testing should cover both alternative encodings and structural reductions.

Diagnosis and repair require distinct evidence. Admission checks the sampled transformation, calibration assesses numerical variation, and aligned comparisons expose inconsistency. None alone guarantees correctness. Canonicalization can remove a discrepancy while leaving the encoding-sensitive operation in place. No tested grid intervention establishes accuracy and feasibility non-inferiority everywhere, and the minimum relation sets cover only the discovered mechanisms. Applying the method to new solvers requires problem-specific reductions, an explicit account of certificate strength, and alignment of random choices for corresponding variables. Future work should validate bounds on the set of optima, test composed relations and unexplained structural failures, and assess whether training from scratch on repaired features preserves consistency and solution quality.

\paragraph*{Data availability}
The \href{\pkglink}{\textcolor{blue}{\textbf{anonymized repository}}} includes relation cards, certificates, a claim ledger, and minimal reproducing examples. The recorded clean build regenerated 165 derived files byte-identically on macOS, with last-digit differences in two tables on Linux\cm{P1}. Third-party code, weights, data, and binaries are not redistributed\cm{P3}. Maintainer reports and six minimal cases are drafted for filing after review\cm{RP7}.

\begin{samepage}
\section*{Acknowledgment}
Generative AI use: Claude Code (Anthropic) was used to draft the text of selected sections, tables, and captions, and wrote code for analysis, experiments, and visualizations. Codex (OpenAI) reviewed all sections and revised wording, figure code, and layout. ChatGPT (OpenAI) audited the study plan. The authors reviewed all text, tested the code, verified all results and references, and take responsibility for the work.
\end{samepage}
\clearpage
\bibliographystyle{IEEEtran}
\bibliography{refs}

\end{document}